\documentclass[10pt,twocolumn,letterpaper]{article}
\pdfoutput=1
\usepackage{iccv}
\usepackage{times}
\usepackage{epsfig}
\usepackage{graphicx}
\usepackage{amsmath}
\usepackage{amssymb}

\usepackage{array}
\usepackage{verbatim}
\usepackage{units}
\usepackage{multirow}
\usepackage{microtype}
\usepackage{booktabs}
\usepackage{amsmath}
\usepackage{amssymb}
\usepackage{graphicx}
\usepackage{bbold}
\usepackage{mathtools}
\usepackage{sistyle}
\usepackage{tabularx}
\usepackage{subcaption}
\SIthousandsep{,}
\usepackage[breakwords]{truncate}

\usepackage[accsupp]{axessibility}

\usepackage[breaklinks=true,bookmarks=false]{hyperref}
\iccvfinalcopy % *** Uncomment this line for the final submission

\def\iccvPaperID{16} % *** Enter the ICCV Paper ID here

\begin{document}

%%%%%%%%% TITLE
\title{Semi-supervised dry herbage mass estimation using automatic data and synthetic images}

\author{Paul Albert$^{1,3,5}$, Mohamed Saadeldin$^{2,3,5}$, Badri Narayanan$^{2,3,5}$, Brian Mac Namee$^{2,3,5}$, \\ Deirdre Hennessy$^{4,5}$, Aisling O{'}Connor$^{4,5}$, Noel O{'}Connor$^{1,3,5}$, Kevin McGuinness$^{1,3,5}$\\\\
$^1$School of Electronic Engineering, Dublin City University\\ $^2$School of Computer Science, University College Dublin\\ $^3$Insight Centre for Data Analytics \\ $^4$Teagasc, $^5$VistaMilk\\
{\tt\small paul.albert@insight-centre.org}
% For a paper whose authors are all at the same institution,
% omit the following lines up until the closing ``}''.
% Additional authors and addresses can be added with ``\and'',
% just like the second author.
% To save space, use either the email address or home page, not both
}
\maketitle
% Remove page # from the first page of camera-ready.

%%%%%%%%% ABSTRACT
\begin{abstract}
Monitoring species-specific dry herbage biomass is an important aspect of pasture-based milk production systems. Being aware of the herbage biomass in the field enables farmers to manage surpluses and deficits in herbage supply, as well as using targeted nitrogen fertilization when necessary. Deep learning for computer vision is a powerful tool in this context as it can accurately estimate the dry biomass of a herbage parcel using images of the grass canopy taken using a portable device. However, the performance of deep learning comes at the cost of an extensive, and in this case destructive, data gathering process. Since accurate species-specific biomass estimation is labor intensive and destructive for the herbage parcel, we propose in this paper to study low supervision approaches to dry biomass estimation using computer vision. Our contributions include: a synthetic data generation algorithm to generate data for a herbage height aware semantic segmentation task, an automatic process to label data using semantic segmentation maps, and a robust regression network trained to predict dry biomass using approximate biomass labels and a small trusted dataset with gold standard labels. We design our approach on a herbage mass estimation dataset collected in Ireland and also report state-of-the-art results on the publicly released Grass-Clover biomass estimation dataset from Denmark. Our code is available at \url{https://git.io/J0L2a}.
\end{abstract}

%%%%%%%%% BODY TEXT
\section{Introduction}
Local monitoring of the biomass composition of  grassland has great potential to improve the reasonable use of fertilizers on dairy farms. Nitrogen over-fertilization has detrimental effects on the environment such as the pollution of underground water or nearby rivers and a reduction in crop yield~\cite{2016_SH_overfertreview,2004_AMBIO_nitrogenchina,2015_JSSPN_fertcentralchile}. Clover is an important ally to reduce the need for nitrogen fertilization as it influences the impact of the fertilization process~\cite{2009_AAS_nitrogen,2009_JAE_biomassMixture}. Mapping the density of the clover content in grassland enables a targeted fertilization (as opposed to a uniform fertilization), which allows farmers to anticipate the amount of nitrogen required, and to limit over-fertilization. Balanced amounts of clover also have a important role to play in the final dry feed for the cow, as sufficient amounts of clover in the cow feed increases food intake and augments dairy production~\cite{2017_JOI_biomassMixture2}.

\begin{figure}[t]
\centering{}\includegraphics[width=\linewidth]{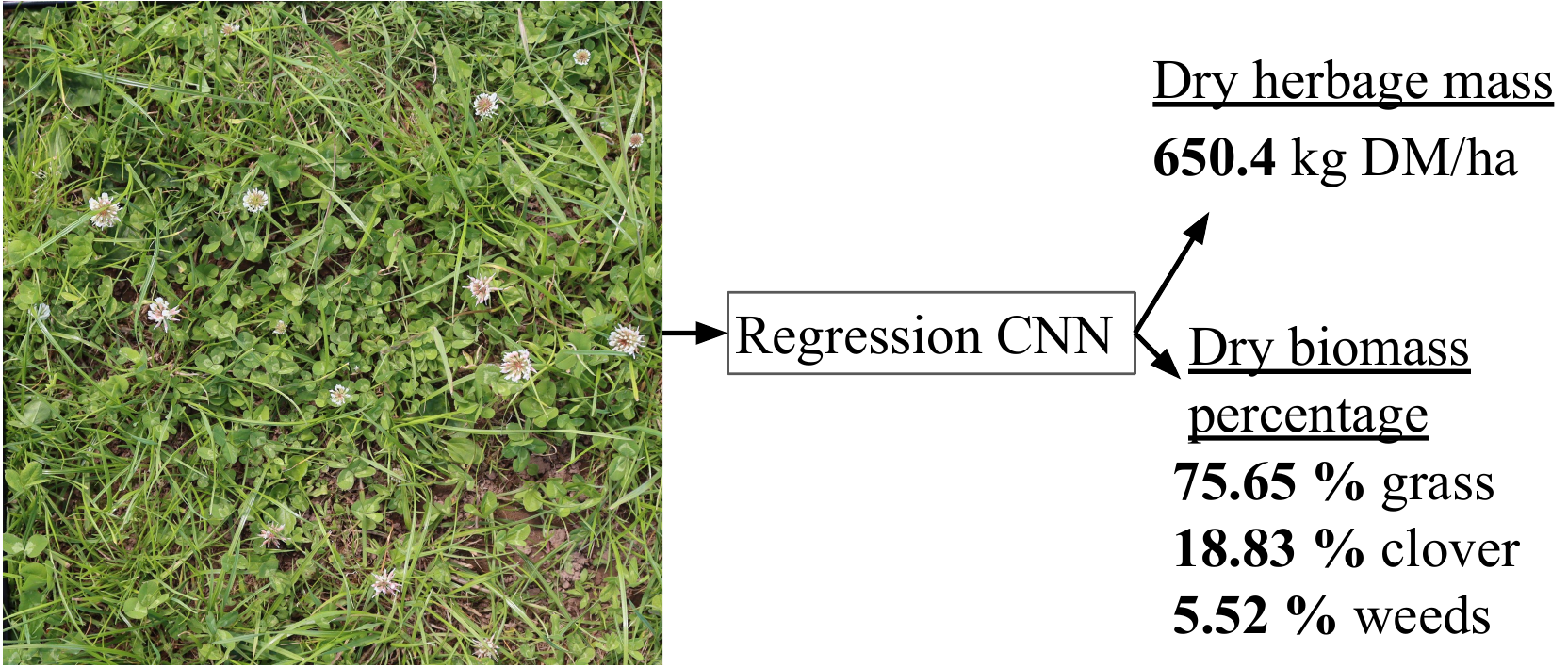}
\caption{Overview of the dry herbage mass prediction task~\label{fig:overview}}
\end{figure}

Species phenotyping proposes a direct application of computer vision where a canopy view of the objects is passed to an algorithm tasked with a computer vision problem. Some examples of these tasks include semantic segmentation~\cite{2019_CVPRW_grasscloverdataset,2014_ECCV_weedsegmentationdataset,2018_ICRA_realtimeseg}, object counting~\cite{2021_arXiv_wheatnet,2020_CVPPP_plantorgancounting}, classification~\cite{2020_PP_globalwheatdataset,2017_arxiv_seedlingclassification,2016_PRL_finelygraineddatasets}, object detection~\cite{2019_PM_deepseedling,2016_Sensors_deepfruits} and regression~\cite{2017_JOI_biomassMixture2,2020_IMVIP_extracting}. The principal limitation when applying deep learning approaches to species phenotyping remains the large amount of annotated data required. Lower supervision alternatives using semi-supervised or unsupervised approaches can lower the annotation burden and enable a stronger convergence than using a small number of annotated images alone. In the case of grass/clover biomass estimation this is even more important, as the annotation process is destructive. To accurately measure biomass the region of interest has to be cut, separated, and weighed in a laboratory whereas the collection of un-annotated images is fast and simple.

In this paper, we use a large collection of unlabeled images together with a small annotated subset to improve the accuracy of a dry herbage mass predicting convolutional neural network (CNN, see Figure~\ref{fig:overview}). We first learn a weakly-supervised semantic segmentation network on synthetic images to estimate the species density in the herbage. We then use the segmentation masks to generate automatic biomass labels for the unlabeled images using a simple regression algorithm. Finally, we train a convolutional neural network on a mix of the automatically labeled data and a small number of manually labeled examples to improve the regression accuracy over training on the small number of manually labeled examples alone. We construct our algorithm on an Irish dry herbage mass dataset~\cite{2021_EGF_Irishdataset} and validate our results on a publicly available dry biomass dataset~\cite{2019_CVPRW_grasscloverdataset} collected in Denmark.

Our contributions are:
\begin{enumerate}
    \item A herbage height aware, weakly supervised, semantic segmentation algorithm trained on synthetic images that is used to automatically label data;
    \item An algorithm leveraging automatically labeled images to improve grass/clover/weed dry herbage mass estimation;
    \item A detailed study of the importance of the low supervision elements for the final accuracy of our algorithm, and a comparison against the state-of-the-art on a publicly available dataset.
    %\item An online challenge with open-source data to be used for the further development of low supervision dry herbage mass estimation
\end{enumerate}
%-------------------------------------------------------------------------
\section{Related work}
 
\subsection{Image analysis for plant phenotyping}
Plant phenotyping and dry matter prediction are excellent domains for the application for image analysis approaches since they enable insight to be extracted from the environment in a non-destructive manner. Existing works explore a variety of computer vision applications for plant phenotyping and in this section we review some of the most relevant to our work. Weed detection aims at localizing unwanted weeds to ultimately remove them by hand or using a robot. Common approaches include employing color filtering, edge detection, and area classification~\cite{2016_RFIUA_weeddetectionsoil,2016_IRJET_weeddetection,2016_CEA_weeddetectchina}; utilising color features used to train random forest algorithms and support vector machines \cite{2020_KSKD_weeddetectchiliRF, 2021_Argriculture_weeddetectUAV}, or using neural networks used to semantically segment images \cite{2017_ICICTI_weedsegcnn}. 
Fruit or vegetable detection and counting reduces human labor by enabling automatic fruit treatment or collection on the farm. Examples include tomato segmentation and counting using a convolutional neural network~\cite{2020_FPS_tomato}, large scale fruit detection in trees~\cite{2016_Sensors_deepfruits}, or real-time detection using a lightweight neural network~\cite{2019_FPS_singleshotpear}.

Some approaches use UAV imaging as opposed to ground-level image capture, introducing a fast solution to mapping weeds in a field~\cite{2019_AAGSSAOP_weedsuav,2021_Argriculture_weeddetectUAV}. As well as using RGB images alone, additional sensors can be added to reduce the difficulty of the phenotyping task~\cite{2010_ISR_differences} such as radar or lidar~\cite{2002_BS_lidar}. 

\subsection{Species biomass estimation from canopy view images}
Biomass estimation from canopy view images aims at providing solutions for targeted fertilization in fields. This opens the way for automated fertilization, reducing costs for the farmer and reducing water pollution due to over-fertilization~\cite{2016_SH_overfertreview,2009_AAS_nitrogen}. The heavy occlusions present in canopy images (see Figure~\ref{fig:overview}) poses significant challenges as the biomass estimate should account for elements hidden from the canopy view.

Himstedt~\etal~\cite{2012_CS_legumegrass} study the biomass of clover in a legume-grass mixture and demonstrate a good capacity to detect clover from the legumes using morphological filtering and color segmentation to detect the clover. The authors were then able to accurately predict the clover biomass in a controlled environment under the assumption that the total biomass is known. Mortensen~\etal~\cite{2017_JOI_biomassMixture2} propose segment the grass clover mixture using color filtering and edge detection before employing a linear regressor to learn the mapping between coverage area of each species and dry biomass content. The authors were then able to directly predict the dry biomass of each element from the image alone. The cow feed scenario presents the added constraint of estimating dry biomass from an image of the fresh pasture.

Skovesen~\etal~\cite{2018_ICPA_grasscloverfirst} propose an improvement over previous work by using a neural network to segment images, and then fitting a linear regressor to the detected species percentages to predict the biomass percentages. To train the neural network, a synthetic dataset is generated using sample crops of relevant species pasted on a soil background. This allows the authors to generate an infinite amount of training images with ground truth from a similar visual domain for their segmentation algorithm. Based on this work, the GrassClover dataset challenge~\cite{2019_CVPRW_grasscloverdataset} asks entrants to improve the author's baseline using the synthetic images together with a large collection of unlabeled real images and a small set of manually labeled real images.

\subsection{Semantic segmentation on synthetic images and domain adaptation}
Semantic segmentation aims at predicting the object that each pixel in an image belongs to~\cite{2011_PASMCL_pascalVOC}. The human annotation required for semantic segmentation tasks is extensive, often requiring several hours per image~\cite{2014_ECCV_MSCOCO}. This makes training strategies using fewer human annotated images attractive. Synthetic images promise to solve part of the problem by providing an unlimited amount of perfectly segmented training images. Popular synthetic datasets for semantic segmentation include The Grand Theft Auto V (GTA V)~\cite{2016_ECCV_GTAV} or SYNTHIA~\cite{2016_CVPR_synthia}  datasets that create synthetic images of cities using graphics engines. 

Although the large quantity of labeled data allows a semantic segmentation neural network to converge on a synthetic dataset, the results need to generalize to real world data. Domain adaptation aims at learning domain agnostic features that can generalize from synthetic data to the real world. Domain adaptation strategies can be applied at different stages in a network: input adaptation, feature adaptation, or output adaptation. Input adaptation strategies aim at transforming synthetic images to look more realistic by applying a realistic style on synthetic images, often using a Generative Adversarial Network~\cite{2017_ICCV_cyclegan,2017_ICCV_gantrans,2017_arXiv_gantransarxiv,2019_CVPR_synsemsegtrans}. 

Feature adaptation approaches aim to discover domain invariant (or aligned) features.  Chen~\etal~\cite{2019_ICCV_maxsquareloss} propose to use a maximum square loss to enforce a linear gradient increase between easier and harder classes. Luo~\etal~\cite{2019_ICCV_adversarialbottleneck} use a significance aware adversarial information bottleneck; Chen~\etal~\cite{2018_CVPR_road} propose a knowledge distillation approach by matching network activations to a network pretrained on ImageNet. 

Output adaptation techniques constrain the network prediction directly to enforce better generalization. This can be achieved using adversarial approaches where the predictions made on synthetic and real data should be indistinguishable to a discriminator network~\cite{2019_CVPRW_outputdiscri}, or by enforcing low entropy (more confident) predictions~\cite{2019_CVPR_adversarialentro}. Batch normalization fine tuning on real data where the batch normalization parameters are tuned on the real images before evaluation has also been shown to be a simple but effective domain transfer strategy~\cite{2017_ICLRW_BNtuning}.
For a more detailed study of domain adaptation for semantic segmentation, we refer the reader to the domain review of  Toldo~\etal~\cite{2020_tech_domainadaptsemsegsurvey}

\subsection{Semi-supervised learning and label noise}
%\ctB{Not sure how much of this is directly relevant to the approach, some could be removed for better clarity}
Training computer vision algorithms with limited supervision aims at learning representative features for a downstream task with little to no supervision. In the scope of this paper we train models using a small annotated subset together with a large amount of un-annotated images, which we refer to as a semi-supervised learning scenario. We additionally introduce label noise literature references, which tackles the scenario of approximately labeled data. This is related to the automatic labels we use in this paper.

%\paragraph{Self-supervised representation learning} learns features from the data itself without the need for labels. The pretext task solved by the neural network on the data is created from the data itself with no need for man-made labels. Examples of pretext tasks include: puzzle solving~\cite{2016_ECCV_Jigsaw}, colorization~\cite{2016_ECCV_Colorful} or rotation prediction~\cite{2018_ICLR_Rotation}. An alternative strategy for self-supervised representation learning gaining traction in the recent literature is unsupervised contrastive learning where the network learns to bring similar images closer together in the feature space while separating dissimilar images. Since learning the features has to be unsupervised, the positive (similar) sample is an augmented version of the initial sample and all other images considered in the batch size are negatives (dissimilar).
%The negative samples can then either be drawn randomly from the distribution~\cite{2020_ICML_SimCLR} or be hard mined using a memory bank~\cite{2020_arXiv_MoCoV2}.

\paragraph{Semi-supervised learning} aims at learning robust features to solve a task using limited annotations. Annotations are necessary in supervised learning to compute the weights of a neural network using gradient descent on a loss computed using the ground truth annotations. In the case of semi-supervised learning, only part of the dataset has been annotated by humans and the rest is unlabeled images. Iteratively approximating labels for the unlabeled data is a tedious task as the errors made by the network will be amplified (confirmation bias~\cite{2020_IJCNN_Pseudo}). State-of-the-art semi-supervised learning uses consistency regularization mechanisms where labels are guessed using multiple views of a sample (different data augmentations)~\cite{2020_ICLR_ReMixMatch}, sometimes coupled with pseudo-labeling~\cite{2020_arXiv_FixMatch}.

\paragraph{Label noise} proposes robust algorithms to mitigate approximate labeling. Approximate labelling can occur when a dataset is created from web queries~\cite{2017_arXiv_WebVision} or when labels are inferred using label propagation~\cite{2021_IJCNN_ReLaB}. Solutions for training a neural network on label noise datasets include lowering the contribution of noisy labels in the training loss~\cite{2020_ICML_MentorMix}, correcting the label using the network prediction~\cite{2019_ICML_BynamicBootstrapping}, meta-learning inspired corrections~\cite{2020_arXiv_MetaSoftApple}, monitoring feature space consistency~\cite{2020_ICPR_Robust}, or robust data augmentation~\cite{2018_ICLR_mixup}.

\section{Biomass prediction in grass-clover pastures}
This section introduces the semi-supervised learning problem of dry biomass estimation of grass-clover pastures, the datasets used, the synthetic image generation process, the automatic labelling pipeline, and our automatic label robust biomass regression algorithm.

\begin{figure*}[t]
\centering{}\includegraphics[width=\linewidth]{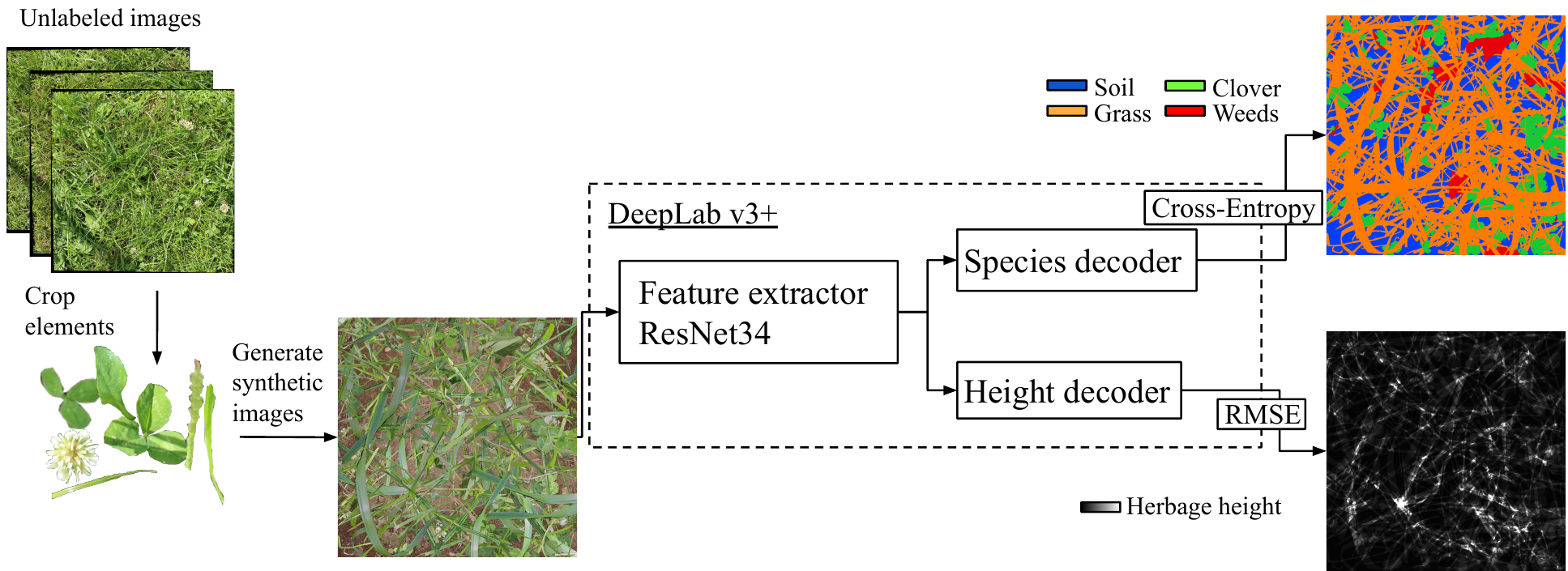}
\caption{Herbage height aware semantic segmentation on synthetic images~\label{fig:synsemseg}}
\end{figure*}

\subsection{Semi-supervised biomass estimation in grass-clover pastures}
We consider here a semi-supervised regression problem with $X_L = \{x_i\}_{i=1}^L$ labeled canopy images of grass and clover, and their corresponding label assignment $Y = \{y_i\}_{i=1}^L,  Y \in \mathbb{R}^C$ where $C$ is the number of species to predict. The small labeled set is complemented by a large set of unlabeled images $X_U = \{x_i\}_{i=1}^D$ with no corresponding labels and $|X_U| \gg |X_L|$. We note the complete dataset used to train the network $X = X_L \cup X_U$. This paper aims to solve the dry biomass prediction problem from images using a convolutional neural network $\Phi: X \rightarrow Y$ using unlabeled images to the improve the regression accuracy.

\subsection{Grass clover dry biomass datasets}
We consider two different dry biomass prediction datasets, both centered around grass and clover biomass prediction. The first dataset we will refer to as the Irish dataset~\cite{2021_EGF_Irishdataset} consists of $528$ images labeled with: total dry herbage mass (kg DM/ha), dry grass biomass percentage (\%), dry clover biomass percentage (\%), and dry weed biomass percentage (\%). We study here the low supervision version of the dataset which includes $156$ fully annotated images ($52$ for training and $104$ for validation) and an additional $594$ unlabeled images. The images were collected using a high resolution Canon camera in Ireland in the summer of 2020. 

The second dataset is the GrassClover dataset~\cite{2019_CVPRW_grasscloverdataset} which contains $152$ images labeled with: dry grass biomass percentage (\%), dry white clover biomass percentage (\%), dry red clover biomass percentage (\%), and dry weed biomass percentage (\%). Contrary to the Irish dataset, the GrassClover dataset distinguishes between red and white clover species but does not target the direct estimation of the dry herbage biomass (kg DM/ha). The fully annotated images are completed with $31\,600$ unlabeled images without corresponding ground truth. The dataset was collected in Denmark in 2017 and 2018. 
%\ctB{Add something here about the phone images in the Irish dataset if we want to do something with it in this paper (generalization to edge devices)}

\begin{figure}[b]
\centering{}\includegraphics[width=1\columnwidth]{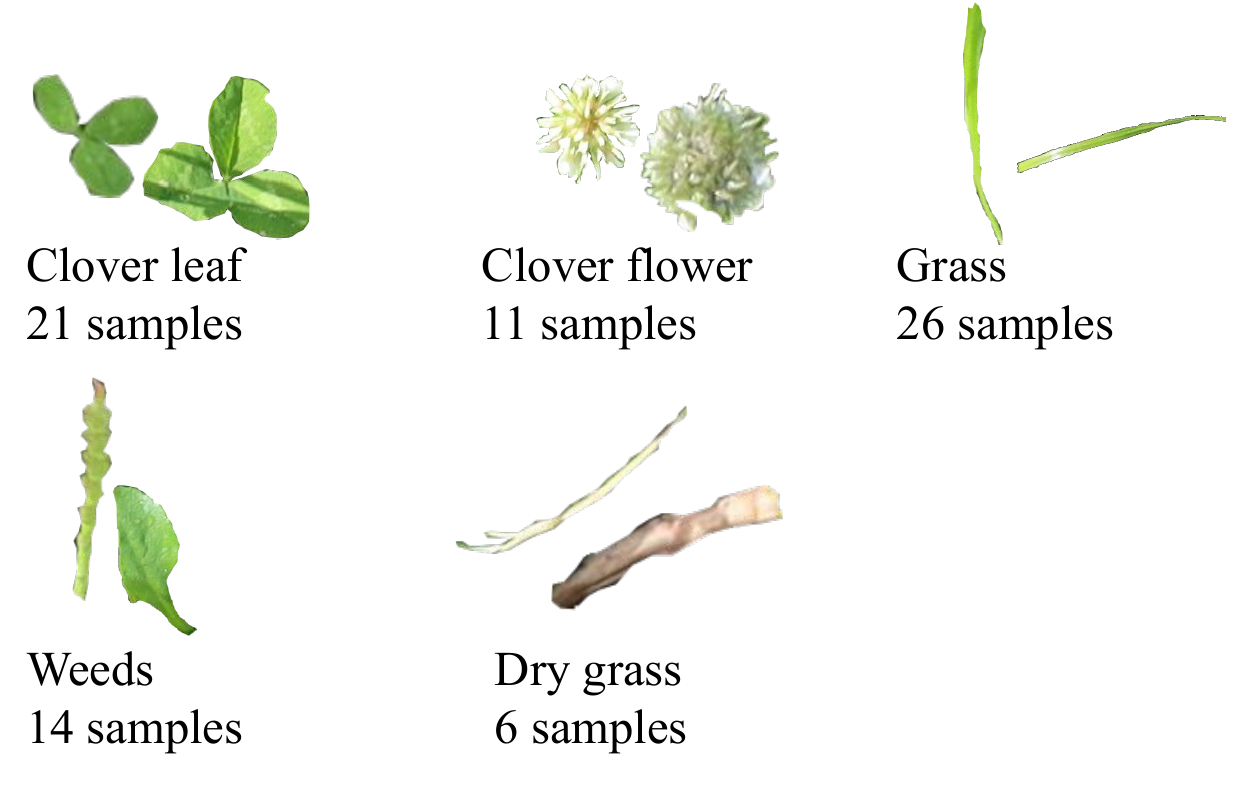}
\caption{Cropped out samples for every species \label{fig:croppedoutsamples}}
\end{figure}

\subsection{Herbage height aware semantic segmentation on synthetic images~\label{sec:automatictheo}}
%To reduce the complexity of directly predicting biomass values from high resolution RGB images, we propose here to simplify the information contained in the herbage images by predicting the observed proportion of each species in the herbage canopy using a semantic segmentation algorithm.
The task we aim to solve in this section is to first predict a semantic segmentation of the herbage into grass, clover (possibly red-white), and weeds; and second, a herbage height map. Since human annotation of ground truth for semantic segmentation  can take up to several hours per image~\cite{2014_ECCV_MSCOCO} and since a pixel specific herbage height is difficult to estimate in practice, we propose (similar to~\cite{2019_CVPRW_grasscloverdataset}) to train our semantic segmentation network $\Psi$ on a synthetically generated dataset $\tilde X$.
We generate the synthetic semantic segmentation images together with their 100\% pixel-accurate synthetic segmentation ground truth using manually cropped out elements from the unlabeled images. In accordance to the low supervision scope of this paper, we only crop out $78$ samples (see Figure~\ref{fig:croppedoutsamples}) and collect $8$ bare soil images to paste elements onto. The bare soil images are collected at the same site and using the same equipment as Hennessey et al.~\cite{2021_EGF_Irishdataset} during the Summer of 2021.

\begin{figure*}[t]
\centering{}\includegraphics[width=.9\linewidth]{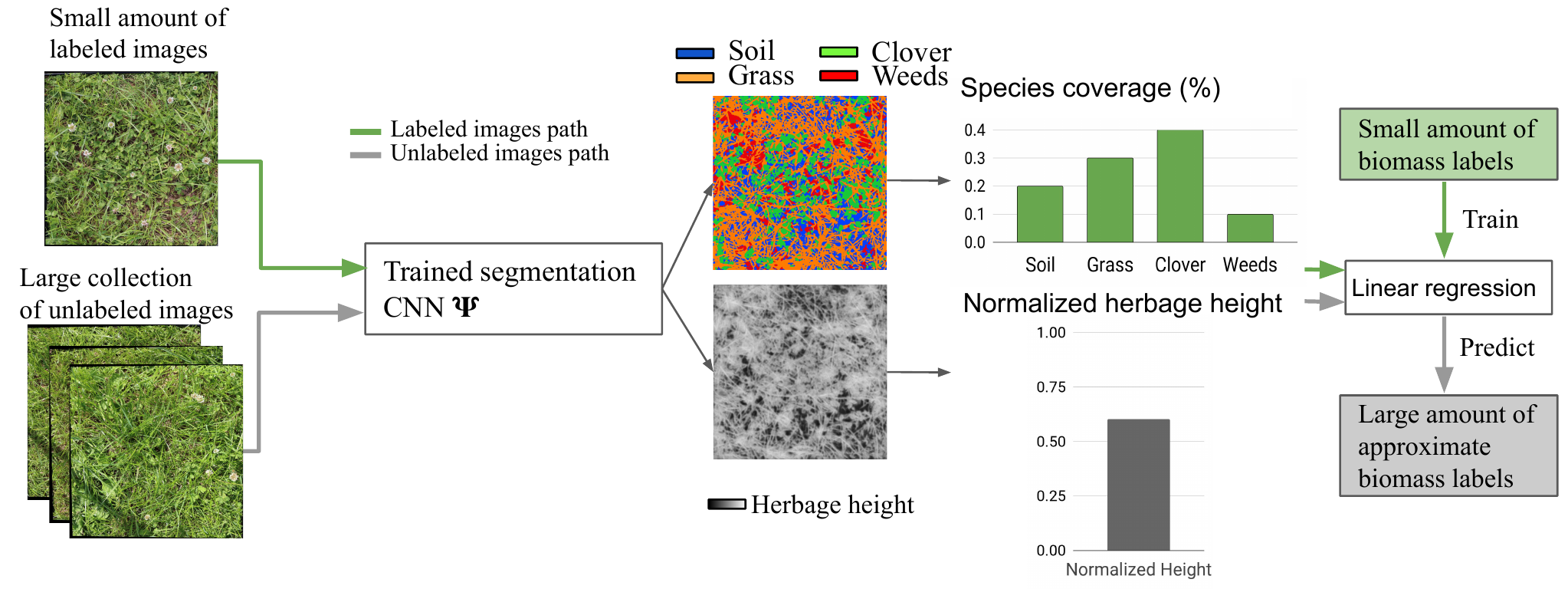}
\caption{Automatic labeling from semantic segmentation \label{fig:regsemseg}}
\end{figure*}

To produce images similar to the real images we aim to make predictions for, we respect the species ratio in images by enforcing the probability of a species to be pasted according to the observed average dry biomass distribution in the training dataset: $90\%$ grass, $7\%$ clover, $3\%$ weeds. We draw the probability of each species to be pasted from a $3$ component Dirichlet distribution with parameters $(9,2,1)$ for (grass, clover, weeds). Once the species has been decided, we randomly draw a sample for this category and apply a series of transformation to increase the diversity of the synthetic images. The transformations include: (uniform) random rotation $(\pm 180^{\circ})$, random Gaussian blur $(\text{radius} \in [0,5])$, random brightness change $[0.6, 1]$, and random resizing $(50-150\%)$. 
Finally, we select a random center location to paste the sample on the background images as well as a mask of the sample's label on the ground truth map.
We additionally approximate the herbage height in the synthetic images as the sum of the total number of successive elements pasted on a pixel. In the rest of the paper, this approximation made on synthetic images will be referred to as herbage height. For example, if three samples have been pasted at the same pixel (clover on top of grass on top of clover), we define the un-normalized herbage height as 3 for the given pixel. Once the synthetic dataset has been fully generated, we compute the 75th percentile of the herbage height for every pixel in all generated images (allowing us to filter outliers) and use this value to clip overly high herbage height numbers and produce a normalized herbage height between 0 and 1 for every pixel in every synthetic image. The normalized herbage height becomes the ground truth target for the segmentation network. Additionally, we found that the quality of the segmentation learnt by $\Psi$ is best when the number of elements to paste is in $[400, 800]$ per image (randomly varied across images); beyond this the synthetic images become overly cluttered. Images are generated at a $2000\times2000$ resolution.
The RGB images are stored in the JPEG format, the grayscale ground truth maps are stored as PNG images, and the herbage height matrix is stored as a compressed numpy array. Figure~\ref{fig:regsemseg} illustrates the automatic labeling pipeline.

%To generate the synthetic dataset, we first crop out examples for each species in the unlabeled images to create a small but diverse collection (see Figure~\ref{fig:croppedoutsamples}). These elements are then randomly pasted on a bare soil background, allowing us to create a large amount of synthetic images with perfect semantic segmentation ground truth labels. Since we aim to predict not only the dry biomass ratios but also the true dry herbage mass on the Irish dataset, we propose to use a herbage height aware segmentation network and concurrently generate a synthetic herbage height map with values in $[0,1]$, linearly proportional to the number of elements pasted on a pixel during the synthetic image creation.
\subsection{Generating synthetic images suitable for herbage mass estimation~\label{sec:synimages}}

To concurrently solve the tasks of semantically segmenting the herbage images and estimating the herbage height for every pixel in the images, we propose a herbage height aware semantic segmentation network $\Psi$ consisting of a single feature extractor coupled with two decoder branches (see Figure~\ref{fig:synsemseg}). 
We concurrently train the species segmentation branch using a pixel-level cross-entropy loss:
\begin{equation*}
\textit{l}_\text{species} = - \sum_{i=1}^C \tilde y_i \log(s_i),
\end{equation*} 
where $S = {s_i}_{c=1}^C$ is the softmaxed prediction of the network and $\tilde{Y} = {\tilde{y}}_{i=1}^C$ are the synthetic segmentation labels. The herbage height branch is trained using a root mean square error (RMSE) loss: 
\begin{equation*}
\textit{l}_{\text{height}} = \sqrt{\frac{1}{P}\sum_{p=1}^P\left(\tilde{h} - h\right)^2}, 
\end{equation*}
where $P$ is the total amount of pixels in the images, $h$ is the ground truth synthetic height label, and $\tilde{h}$ is the network prediction (sigmoid). The total training loss of the segmentation network $\Psi$ is $\mathit{l} = l_\text{specices} + l_\text{height}$.

\begin{table*}[]
    \caption{Importance of data augmentation and batch normalization tuning when training on synthetic images.\label{tab:ablasemsegaug}}
    \global\long\def\arraystretch{1}%
    \centering
    \resizebox{.9\textwidth}{!}{{{}}%
    \begin{tabular}{l>{\centering}c>{\centering}c>{\centering}c>{\centering}c>{\centering}c>{\centering}c>{\centering}c>{\centering}c>{\centering}c>{\centering}c}
    \toprule
        & \multicolumn{5}{c}{HRMSE} & & \multicolumn{4}{c}{RMSE} \tabularnewline
        \cmidrule(lr){2-6}\cmidrule(lr){8-11}   
        & Total & Grass & Clover & Weeds & Avg. & HRAE & Grass & Clover & Weeds & Avg. \tabularnewline
        \midrule
        Simple DA & 357.35 & 328.66 & 55.74 & 26.75 & 137.05 & 35.26 & 8.11 & 6.87 & 3.22 & 6.07 \tabularnewline
        + ColorJitter & 319.92 & 289.32 & 60.81 & 31.40 & 127.18 & 35.46 & 8.63 & 7.68 & 3.55 & 6.62 \tabularnewline
        + BN tuning & 284.60 & 258.34 & 51.92 & 27.05 & 112.44 & 31.79 & 6.49 & 4.94 & 3.24 & 4.89 \tabularnewline
       \bottomrule
    \end{tabular}}
\end{table*}

\subsection{Automatic label prediction from species density estimations}
The herbage height aware semantic segmentation network $\Psi$ allows us to reduce the complexity of the biomass prediction problem by simplifying the input domain from high resolution real RGB images to the surface area occupied by each species in the canopy as well as an estimated herbage height map. From there, we compute the relative area occupied by each species in the canopy (in \%) and the predicted herbage height over each image and train a simple ridge regression algorithm using the small number of labels, $Y$, to predict approximate labels for $X_U$. This intermediate task allows us to generate accurate automatic labels for $X_U$ even if the number of images in $X_L$ is very limited.

\subsection{Regression on automatic labels with a trusted subset}
Although the biomass information can be directly predicted using the automatic annotation process (as done in Skovsen et al.~\cite{2019_CVPRW_grasscloverdataset}), we propose to attempt to decrease the regression error further by solving the regression problem directly from the RGB images using a convolutional neural network, $\Phi$, and both human-labeled and automatically labeled image datasets: $X_L$ coupled with ground truth labels $Y$ (the trusted set) and $X_U$ coupled with approximate labels $\tilde Y$ (the automatically labeled set). $\Phi$ is trained to predict the biomass composition (\%) and the dry herbage mass (kg DM/ha) from RGB images alone; the automatic images are only used in $\Psi$ to help predict the automatic labels $\tilde Y$ for unlabeled images in $X_U$.
To ensure that $\Phi$ will not over-fit to incorrect approximate labels, we use three mechanisms. First, we over-sample the trusted data to ensure that a fixed percentage will always be presented to the network in every mini-batch ($\frac{3}{4}$ approximate labels, $\frac{1}{4}$ trusted labels). Second, we use a label perturbation strategy where we randomly perturb the automatic labels to avoid over-fitting incorrect targets, and to avoid penalizing the network for making a prediction slightly different than the incorrect prediction. In practice, we randomly perturb the label by $\pm$ two times the observed RMSE of the automatic labels on the validation set. Finally, we find that adding vertical flipping and randomly grayscaling to the input images to be interesting augmentations that preserve the full herbage information of the image and help further decrease validation error.
\begin{table*}[]
    \caption{Ablation study for predicting approximate labels. We report the biomass prediction errors on a heldout validation set. \textbf{HL}: hard labels, \textbf{SL}: soft labels, \textbf{H}: herbage height \label{tab:ablaapprox}}
    \global\long\def\arraystretch{1}%
    \centering
    \resizebox{.9\textwidth}{!}{{{}}%
    \begin{tabular}{l>{\centering}c>{\centering}c>{\centering}c>{\centering}c>{\centering}c>{\centering}c>{\centering}c>{\centering}c>{\centering}c>{\centering}c}
    \toprule
        & \multicolumn{5}{c}{HRMSE} & & \multicolumn{4}{c}{RMSE} \tabularnewline
        \cmidrule(lr){2-6}\cmidrule(lr){8-11}   
        & Total & Grass & Clover & Weeds & Avg. & HRAE & Grass & Clover & Weeds & Avg. \tabularnewline
        \midrule
        HL & 351.54 & 332.88 & 51.34 & 28.29 & 137.50 & 41.61 & 6.82 & 6.20 & 3.25 & 5.42 \tabularnewline
        SL & 310.68 & 279.98 & 57.48 & 28.15 & 121.87 & 34.18 & 7.61 & 5.20 & 3.24 & 5.35 \tabularnewline        
        HL + SL & 315.20 & 288.52 & 53.37 & 28.11 & 123.33 & 34.33 & 6.49 & 4.91 & 3.23 & 4.88 \tabularnewline
        HL + SL + H & 284.60 & 258.34 & 51.92 & 27.05 & 112.44 & 31.79 & 6.49 & 4.94 & 3.24 & 4.89 \tabularnewline
       \bottomrule
    \end{tabular}}
\end{table*}

\section{Experiments}
\subsection{Training details~\label{sec:traindetails}}
We use two different neural networks to solve two distinct tasks. For the semantic segmentation network $\Psi$, we use a state-of-the-art architecture: DeepLabV3+~\cite{2018_ECCV_DeepLabv3} where we duplicate the decoder to create the herbage height branch. $\Psi$ is trained on $800$ synthetic images and uses $200$ synthetic images for validation. We use a ResNet34~\cite{2016_CVPR_ResNet} as the feature extractor, initialized on ImageNet~\cite{2012_NeurIPS_ImageNet}, and with an output stride of 16 for both training and testing. We use the ``poly" lr schedule~\cite{2018_ECCV_DeepLabv3} starting at $0.007$, a batch size of $4$, and train for $60$ epochs. For the base data augmentation we resize images to $1024$ on the short size, randomly crop a $1024\times1024$ square, randomly flip horizontally, and normalize the images.

For the regression network $\Phi$, we use a ResNet18 network~\cite{2016_arXiv_Wide} pretrained on ImageNet to solve the regression problem from RGB images directly. We train for 100 epochs, starting with a learning rate of $0.03$ dividing it by 2 at epochs $50$ and $80$. We use the same base data augmentation as for $\Psi$ but with a resolution lowered to $512\times512$. For the strong(er) data augmentation, we add random vertical flipping and random grayscaling ($p=0.2$). We train with a batch size of $12$.

We use the Irish dataset~\cite{2021_EGF_Irishdataset} in its low supervision configuration ($52$ images are used for training, $104$ for validation and $372$ for testing) for our exploratory studies, and generate 1000 synthetic images to train $\Psi$ according to the process described in Section~\ref{sec:synimages}. We validate our results on the GrassClover dataset~\cite{2019_CVPRW_grasscloverdataset} and use the full $152$ fully annotated biomass images, dividing them into $100$ for training and $52$ for validation; we use the $174$ images withheld for the CodaLab~\footnote{https://competitions.codalab.org/competitions/21122} for testing. We make use of $800$ randomly selected synthetic images out of the $8000$ generated by the authors for $\Psi$, keeping $200$ extra images for validation. We do not train the herbage height branch on the GrassClover dataset.

To evaluate the performance of the algorithms, we report the RMSE when predicting the dry biomass species percentage for both the Irish and GrassClover datasets. For the Irish dataset, we additionally report the RMSE of the global herbage mass prediction (HRMSE, kg DM/ha), the herbage relative absolute error $l_\text{relative} = \frac{1}{N}\sum_{i=1}^N \frac{|y_i-\tilde{y_i}|}{y_i}$ (HRAE, in \%) and the HRMSE specific to each species (kg DM/ha).

\subsection{Semantic segmentation on synthetic images}
To encourage $\Psi$ to learn robust features that will generalize to unseen real images, we augment the synthetic images using color jittering and Gaussian blur. Furthermore, once the network has converged on the synthetic dataset and before predicting on the real images, we perform batch normalization tuning which is a common domain adaptation strategy~\cite{2017_ICLRW_BNtuning} on the real images. An ablation study on the importance of the data augmentation and batch normalization tuning is given in Table~\ref{tab:ablasemsegaug}, where we use the best performing regression algorithm from~\ref{sec:regexp}.
\begin{table*}[]
    \caption{Ablation study on training with approximate labels. We report results on the validation set using the linear regression baseline \textbf{LR} or training on the trusted data only \textbf{T}, the automatic data only \textbf{A}, or combinations of both \textbf{T+A} \label{tab:ablaapproxcnn}}
    \global\long\def\arraystretch{1}%
    \centering
    \resizebox{.95\textwidth}{!}{{{}}%
    \begin{tabular}{l>{\centering}c>{\centering}c>{\centering}c>{\centering}c>{\centering}c>{\centering}c>{\centering}c>{\centering}c>{\centering}c>{\centering}c}
    \toprule
        & \multicolumn{5}{c}{HRMSE} & & \multicolumn{4}{c}{RMSE} \tabularnewline
        \cmidrule(lr){2-6}\cmidrule(lr){8-11}   
        & Total & Grass & Clover & Weeds & Avg. & HRAE & Grass & Clover & Weeds & Avg. \tabularnewline
        \midrule
        LR & 284.60 & 258.34 & 51.92 & 27.05 & 112.44 & 31.79 & 6.49 & 4.94 & 3.24 & 4.89 \tabularnewline
        T & 249.48 & 253.63 & 45.62 & 32.67 & 110.64 & 21.67 & 6.28 & 5.07 & 3.94 & 5.10 \tabularnewline
        A & 258.00 & 239.81 & 46.51 & 27.74 & 104.69 & 23.48 & 5.72 & 5.20 & 3.29 & 4.74 \tabularnewline
        T + A & 245.04 & 233.34 & 34.94 & 26.32 & 98.20 & 21.60 & 4.70 & 4.45 & 3.17 & 4.11 \tabularnewline
        + random GS & 234.25 & 217.55 & 37.57 & 27.72 & 94.28 & 21.55 & 4.66 & 4.47 & 3.27 & 4.13 \tabularnewline
        + trusted oversampling & 232.08 & 220.09 & 35.93 & 26.34 & 94.12 & 21.36 & 4.33 & 4.17 & 3.15 & 3.88 \tabularnewline
        + random perturbation & 229.93 & 216.23 & 35.79 & 26.05 & 92.69 & 19.96 & 4.22 & 4.21 & 3.10 & 3.84 \tabularnewline
       \bottomrule
    \end{tabular}}
\end{table*}

\begin{table*}[]
    \caption{Results on the GrassClover test set (RMSE).\label{tab:resultsdanish}}
    \global\long\def\arraystretch{1}%
    \centering
    \resizebox{.65\textwidth}{!}{{{}}%
    \begin{tabular}{l>{\centering}c>{\centering}c>{\centering}c>{\centering}c>{\centering}c>{\centering}c>{\centering}c>{\centering}c}
    \toprule
        &  &  \multicolumn{3}{c}{Clover} & &  \tabularnewline
        \cmidrule(lr){3-5}
        & Grass & Total & White & Red & Weeds & Avg. \tabularnewline
        \midrule
        Skovsen~\etal~\cite{2019_CVPRW_grasscloverdataset} & 9.05 & 9.91 & 9.51 & 6.68 & 6.50 & 8.33 \tabularnewline
        Naranayan~\etal~\cite{2020_IMVIP_extracting} & 8.64 & 8.73 & 8.16 & 10.11 & 6.95 & 8.52 \tabularnewline
        \midrule
        Trusted data & 10.28 & 10.32 & 9.24 & 9.54 & 7.37 & 9.35 \tabularnewline
        + Automatic data & 8.78 & 8.35 & 7.72 & 7.35 & 7.17 & 7.87 \tabularnewline
       \bottomrule
    \end{tabular}}
\end{table*}

\subsection{Regression from species coverage~\label{sec:regexp}}
We compare different sets of simple features to extract from the segmentation masks as well as the importance of the herbage height prediction when estimating the dry herbage mass. For features directly related to the dry biomass percentages, we compare averaging the most confident prediction for every pixel only (hard label, HL), averaging the full softmax prediction at each pixel (soft label, SL), or using the two sets of features jointly (HL+SL). In the regression model each feature is the average of the observations over the whole image: 4 features (soil \%, grass \%, clover \%, weeds \%) for HL or SL (8 for HL+SL), and 1 feature for the herbage height.

We fit a least squares $L_2$ regularized (ridge) regression algorithm to all features with a regularization factor of $1$, and train on the small subset of annotated images before evaluating on the validation set (Table~\ref{tab:ablaapprox}). 
First, we report the RMSE error of the total herbage mass error (kg DM/ha), as well as the detailed grass/clover/weed herbage mass estimation (kg DM/ha). Second, we report the relative RMSE for the total herbage mass (\%) and the RMSE for the relative dry biomass estimation (\%) for the grass/clover/weeds.
We notice that using SL is better than HL when predicting the herbage mass, demonstrating the interest of capturing the full softmax information over the max prediction only. The precision of  HL is still beneficial as we observe a good improvement in terms of dry biomass percentage RMSE when the two sets of features are coupled. When adding the information about the herbage height, a  decrease in HRMSE error is observed, validating the importance of the herbage height module in the segmentation architecture.

\subsection{Biomass prediction using automatic labels and a trusted subset}
We use the automatic labels to enhance the generalization of the regression CNN $\Phi$ in order to improve over the linear regression from the predictions of $\Psi$, especially in terms of herbage mass prediction.
Table~\ref{tab:ablaapproxcnn} reports the ablation study showing how the additional mechanisms we introduce allow us to be robust to the approximate automatic regression labels. The reported metrics are described in Section~\ref{sec:traindetails}. We also compare the performance of the regression network against the linear regression from the prediction of $\Psi$.

\subsection{Comparison against other works on the GrassClover dataset~\label{sec:compdanish}}
We compare the improvements of our approach on the publicly released GrassClover dataset~\cite{2019_CVPRW_grasscloverdataset}. The target metrics for this dataset are limited to the dry biomass percentages, for which we report RMSE errors. Table~\ref{tab:resultsdanish} reports the performance  of our algorithm with and without automatic labels on the test set available on the CodaLab challenge~\footnote{https://competitions.codalab.org/competitions/21122} and compares against the best available results.
We report a lower RMSE on average than the methods we compare against and show that our algorithm is capable of using unlabeled images to reduce the biomass estimation error for every species over training on the small trusted subset alone.
%\ctB{I would add here another table with our results on the held out Irish test set which would be the comparison point of other algorithms if we are allowed to release a public challenge}

\section{Conclusion}
This paper proposes to improve upon existing low supervision baselines in dry grass clover biomass prediction by making use of unlabeled images. To do so, we first train a herbage height aware semantic segmentation network on synthetic images that we use to generate automatic labels for the unlabeled data using a small set of labeled images. We then train a regression CNN on RGB images directly using the automatic labels to improve the accuracy over using the trusted data alone. We demonstrate the importance of our herbage height aware segmentation network when predicting dry herbage masses from canopy view images as well as the noise robust mechanisms we use to train on automatically labeled data. We improve over our baseline on the Irish dry herbage biomass dataset and set a new state-of-the-art performance level on the publicly available GrassClover dataset. %\ctB{Maybe add something about interpreting the quality of the results (human error ~20\% on the herbage mass)}

\ificcvfinal\section*{Acknowledgments} This publication has emanated from research conducted with the financial support of Science Foundation Ireland (SFI) under grant number SFI/15/SIRG/3283 and SFI/12/RC/2289\_P2.\fi

{\small
\bibliographystyle{ieee_fullname}
\bibliography{egbib}
}

\end{document}